\documentclass[10pt,twocolumn,letterpaper]{article}

\usepackage{cvpr}
\usepackage{times}
\usepackage{epsfig}
\usepackage{graphicx}
\usepackage{amsmath}
\usepackage{amssymb}
\usepackage[utf8]{inputenc} % allow utf-8 input
\usepackage[T1]{fontenc}    % use 8-bit T1 fonts
\usepackage{url}            % simple URL typesetting
\usepackage{booktabs}       % professional-quality tables
\usepackage{amsfonts}       % blackboard math symbols
\usepackage{nicefrac}       % compact symbols for 1/2, etc.
\usepackage{microtype}      % microtypography
\usepackage{algorithm}
\usepackage{graphicx}
\usepackage{wrapfig}
\usepackage{subcaption}
\usepackage[noend]{algpseudocode}
\usepackage{caption}
\usepackage[pagebackref=true,breaklinks=true,letterpaper=true,colorlinks,bookmarks=false]{hyperref}

\cvprfinalcopy% *** Uncomment this line for the final submission

\def\cvprPaperID{2191} % *** Enter the CVPR Paper ID here

\begin{document}

% Draft title
\title{Adaloss: Adaptive Loss Function for Landmark Localization}

% The \author macro works with any number of authors. There are two
% commands used to separate the names and addresses of multiple
% authors: \And and \AND.
%
% Using \And between authors leaves it to LaTeX to determine where to
% break the lines. Using \AND forces a line break at that point. So,
% if LaTeX puts 3 of 4 authors names on the first line, and the last
% on the second line, try using \AND instead of \And before the third
% author name.

\author{
  Brian Teixeira, Birgi Tamersoy, Vivek Singh, Ankur Kapoor \\
  Siemens Healthineers, Digital Services, Digital Technology \& Innovation, \\
  Princeton, NJ, USA
}

\maketitle

\begin{abstract}
%% This paper presents a novel approach to train deep networks for
%% regressing landmarks in images. Existing methods have limitations due
%% to target representations often either regressing landmarks as vector
%% of 2D coordinates or heatmap with Gaussian distribution at the target
%% location. While the former method has difficulty in handling missing
%% landmarks, the later is limitted in the stability of training
%% depending upon the desired target precision. We propose a method to
%% address the limitation of heatmap regression method by adapting the
%% target loss function during training by iteratively adjusting the
%% variance of the target Gaussian. We demonstrate the effectiveness of
%% the proposed method on the challenging tasks of estimating human pose
%% in natural scenes and detecting catheter tip in medical X-ray images.
% \footnote{This feature is based on research, and is not
%  commercially available. Due to regulatory reasons its future
%  availability cannot be guaranteed.}

Landmark localization is a challenging problem in computer vision with a multitude of applications. Recent deep-learning based methods have shown improved results by regressing likelihood maps (i.e. heatmaps) instead of regressing the coordinates directly. However, setting the precision of these regression targets during the training has been a cumbersome process since it creates a trade-off between trainability vs.\ localization accuracy. Using precise targets introduces a significant sampling bias and hence makes the training more difficult, whereas using imprecise targets results in inaccurate landmark detectors. In this paper, we introduce ``Adaloss'', an objective function that adapts itself during the training by updating the target precision based on the training statistics. This approach does not require setting problem-specific parameters and shows improved stability in training and better localization accuracy in inference. We demonstrate the effectiveness of our proposed method in three very different applications of landmark localization: 1) the challenging task of precisely detecting catheter tips in medical X-ray images\footnotemark, 2) localizing surgical instruments in endoscopic images\footnotemark[\value{footnote}], and 3) localizing facial features on in-the-wild images where we show state-of-the-art results on the 300-W benchmark dataset. 

\footnotetext{This feature is based on research and is not commercially available. Due to regulatory reasons its future availability cannot be guaranteed.}

%Landmark detection is a challenging problem in computer vision.
%Existing methods have limitations due to the target representation
%which highly depends on the nature of the problem. While regressing
%probability maps has shown better efficiency for landmark estimation,
%setting the precision of the regression target has been a
%cumbersome process since it creates a trade-off between trainability
%vs.\ localization accuracy. In this
%paper, we introduce ``Adaloss'', an objective function that adapts
%itself during the training by updating target precision
%based on the training statistics. This approach does not require
%setting problem specific parameters and shows improved stability in
%training and inference. We demonstrate the effectiveness of the
%proposed method on the challenging tasks of detecting catheter tip 
%in medical X-ray images \footnote{This feature is based on research, and is not
% commercially available. Due to regulatory reasons its future
% availability cannot be guaranteed.} and localizing keypoints on in-the-wild faces, 
%where we show state of the art results.

\end{abstract}

\section{Introduction}

%Automatic localization of landmarks or semantic keypoints in images is
%a challenging problem with far reaching impact on several domains such
%as human pose estimation \cite{hg, mpii}, facial image analysis
%\cite{hyperface} and medical image analysis \cite{ghesu17,
%  brian_cvpr18}. Recent years have seen tremendous progress with the
%availability of large benchmark datasets such as 300-W \cite{300w},
%MPII \cite{mpii} or FLIC\cite{flic} and improvements in training deep networks
%\cite{hg, insafutdinov16ariv, yang, ke}.

% importance of the problem, selling the "generic" nature...
Landmark localization is a fundamental problem in computer vision with far-reaching impact on a myriad of applications in multiple domains. Whether the goal is to analyze the attentiveness of a driver \cite{jabon2011}, interpret sign language \cite{Grzejszczak2016MTA}, plan aortic valve surgeries \cite{zheng2010}, estimate body pose for human-computer interaction \cite{shotton13pami}, or detect surgical instruments in endoscopic procedures \cite{du2018}, the first step usually involves precise detection of relevant semantic keypoints in the given input images.

\begin{figure}[t!]
\centering
\includegraphics[width=0.8\linewidth]{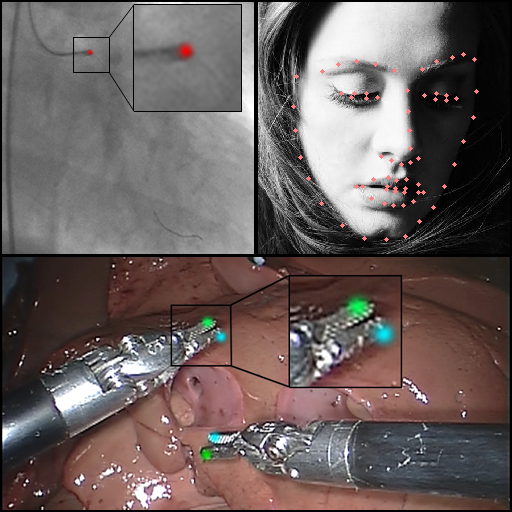}
  \caption{From one task to another, landmark localization has far reaching impact on multiple domains. (top-left) Detection of the tip of a catheter in a medical X-ray image, where the tolerance for practical usefulness is about 1 pixel at 256 x 256 resolution and annotations are very precise. (top-right) Facial feature localization, where some landmarks can be defined very precisely (e.g. eye corners), but others cannot be defined as precisely (e.g. chin landmarks). (bottom) Surgical instrument detection in endoscopic procedures, where the inherent occlusions and the viewpoint make it a challenging task. 
  }
  \label{fig:first}
\end{figure}

% constrain the discussion to DL-era...
Similar to the other fundamental computer vision problems such as object classification and segmentation, landmark localization has also seen a tremendous progress in recent years due to the availability of large datasets (e.g. \cite{300w,du2018,afw,mpii,flic}), improvements in training deep neural networks (e.g. \cite{unet,batchnorm,kingma2015,relu}), and availability of more  computational resources. 

\begin{figure*}[ht!]
  \begin{center}
    \centering
    \includegraphics[width=0.78\textwidth]{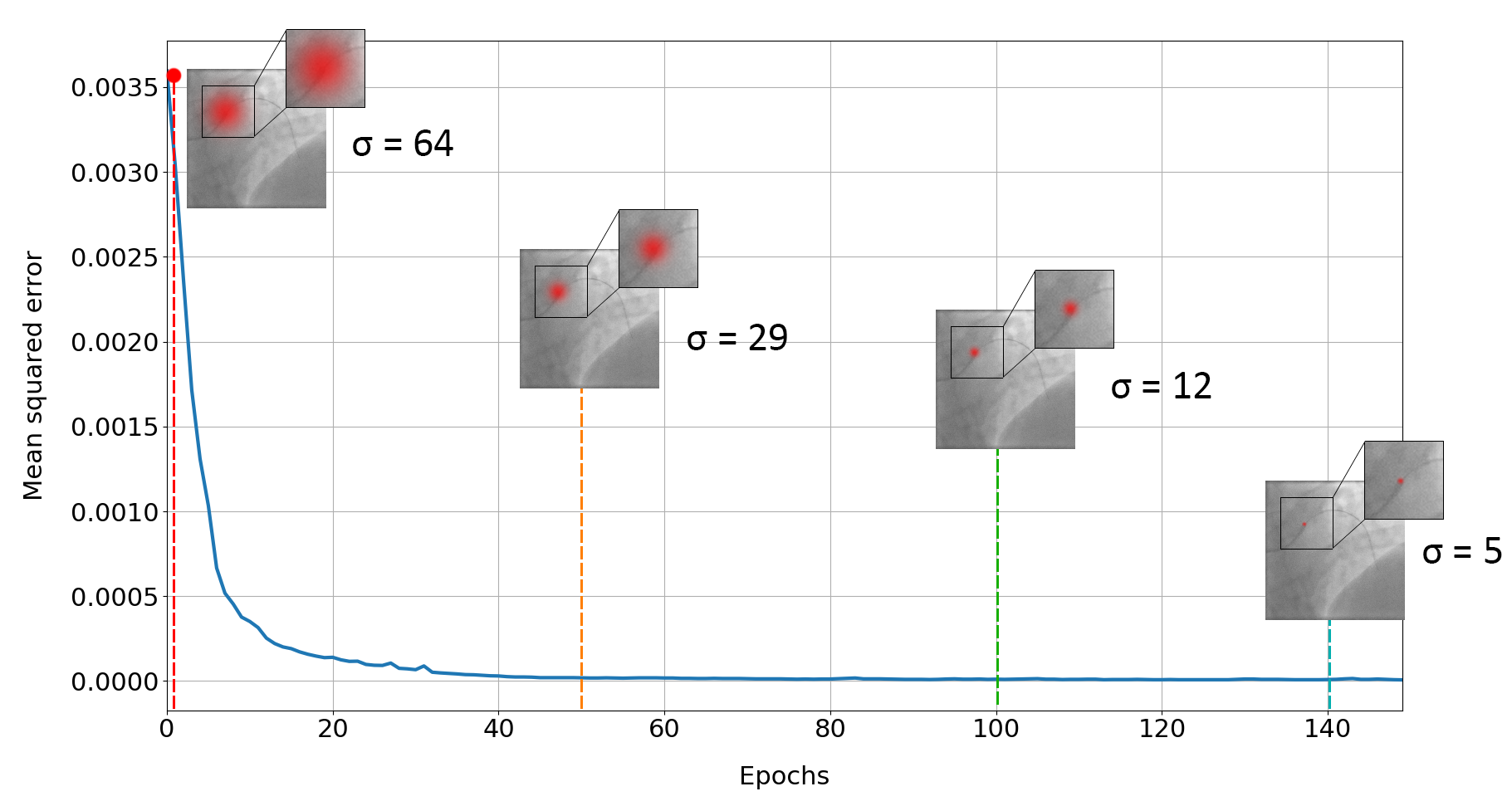}
	\caption{Evolution of Adaloss during training. In the early
          stages, Adaloss uses a large variance for the
          the target heatmaps. As the training progresses, Adaloss
          gradually decreases the target variance until reaching
          the optimal value for the given application.}
	\label{fig:evol_cath_loss}
  \end{center}
\end{figure*}

% explain how existing methods formulate the problem and what are the shortcomings (both for direct coordinate regression & heatmap regression - possibly separately). 
Most existing deep-learning based methods formulate the landmark localization problem as a structured output regression problem where the landmark locations are either represented as a set of image coordinates \cite{zhangpami,wu2017}, or as a set of likelihood maps (i.e. heatmaps) \cite{thompson,hg,honari2018,san,dong2018}. While the former representation certainly provides a compact output, it has several undesirable properties such as: 1) forcing the network to learn the arbitrary mapping between the appearance of a landmark and its image coordinates,  2) outputting a single target location for each landmark instead of a number of candidate locations, and 3) forcing the design of explicit mechanisms for handling missing landmarks. As a result, these networks are usually harder to train and their output is less usable as a pre-processing step of more complex vision tasks. On the other hand, training deep networks to regress likelihood maps as landmark locations does not suffer from these limitations.

%Multitude of existing methods for localizing landmarks formulate it as
%a structured output regression problem, either representing the
%landmarks as a vector of 2D coordinates \cite{zhangpami}, or a
%multi-channel image where each channel is a probability map with a
%gaussian distribution centered at the target landmark location
%\cite{thompson, hg, wei, yang, chu, ke}. While the former approach certainly has a compact output,
%it forces the network to identify a single target location for
%each landmark; since landmark detection often serves as a
%pre-processing for other tasks such as model fitting or tracking, this
%constraint limits the usefulness of the output. Furthermore, for
%scenarios where not all landmarks may be present in every image, such
%networks have to be trained to report null or a default location
%suggesting the absence which makes the training difficult. On the
%other hand, training deep networks to predict the probability map of
%landmarks does not suffer from these limitations. For instance, the
%output can be interpreted as a likelihood map for obtaining multiple
%candidate target locations (\autoref{fig:first}).

% heatmap regression difficulties, emphasize the "problem specific ad-hoc solutions" to the limitations! we also need to mention that existing methods use one single sigma for all landmarks in multi-landmark regression problems!
A ubiquitous way of defining these target likelihood maps in training is creating multi-variate Gaussian distributions centered at the target landmark locations \cite{thompson, hg, honari2018, san,dong2018, wei,yang, chu,ke}. The main concern here is appropriately setting the variance of these Gaussian distributions. Ideally, one would use the smallest possible variance (practically a delta function) while generating these target heatmaps so that the resulting detectors will be as precise and as accurate as possible. However, this does not work in practice. First, there is an inherited uncertainty and inconsistency in the human annotations \cite{lampert2016} that needs to be considered during the training. Furthermore, training deep networks with highly sparse output distributions is difficult. This may be attributed to the significant sample bias in the outputs. With such bias during training, the networks quickly degenerate to all empty outputs and then over the course of the training they attempt to recover peaks around the target locations of various samples. Existing methods try to address these problems by \textit{application-dependent} and \textit{add-hoc} solutions, such as using very low learning rates \cite{hg,unite}, arbitrarily weighting the target heatmaps \cite{davison2018,neil2018}, or inevitably sacrificing the inference accuracy by using Gaussian distributions with larger variances. 

%A primary concern when preparing target heatmaps for regression is to
%appropriately set the variance of the gaussian distribution. Ideally,
%the variance should be set based on the expected error tolerance for the
%subsequent task. For instance, to detect a
%catheter tip (only a few pixels) in an X-ray image (often 512x512 or
%1024x1024 resolution), the variance of gaussian distribution must be
%set to be as small as possible. However, training networks with such
%highly sparse output distribution is difficult. This may be attributed
%to the fact that such output distribution has a significant sample
%bias with only a few output pixels with positive values. With such bias,
%during training the network quickly degenerates to all empty heatmaps
%and over the course of training it attempts to recover peaks around the
%target location for various samples. This limitation often forces use
%of very low learning rates resulting in very long training cycles
%\cite{hg, ke, unite, yang}, or inevitably sacrificing on the
%accuracy by using gaussian distribution with larger variances.

In this paper, we introduce a novel, \textit{application-independent} method for landmark localization, which addresses the aforementioned limitations. Our approach, which we refer to as ``Adaloss'', \textit{indirectly} adapts the objective function throughout the training by updating the training targets (see Figure \ref{fig:evol_cath_loss}). Adaloss begins the training by using targets with large variances, and then iteratively adjusts the target variances based on \textit{landmark-specific} training statistics. Effectively, Adaloss provides an implicit ``curriculum'' for training.

Besides being easy to develop and integrate into existing training pipelines, Adaloss also increases the robustness of the training in applications with highly sparse target distributions, is less sensitive to initial learning rate, and requires fewer iterations to converge. It also generates more stable networks \cite{deformstab} and is capable of producing highly accurate landmark detectors in multiple domains. We demonstrate the effectiveness of our method in three very different applications of landmark localization: 1) detecting catheter tips in X-ray images where a single instance of a single landmark needs to be localized very precisely, 2) facial feature localization where single instances of multiple landmarks need to be localized in varying precision, where we show state-of-the-art results, and 3) detecting surgical instruments in endoscopy images where multiple instances of multiple landmarks need to be localized very precisely.

%In this paper, we introduce a novel method for landmark regression
%which aims to address the aforementioned concerns by progressively
%changing/adapting the objective function; we refer to the proposed method as ``Adaloss''. %Instead of starting the training to regress a
%gaussian distribution with small variance (desired target), our
%approach begins by training the network to regress a gaussian
%distribution with large variance, and then iteratively adjusts the
%variance based on the loss history, eventually ensuring that upon
%convergence, the final trained network would predict gaussian
%distribution with an originally-desired small variance. Besides being
%easy to develop and incorporate into existing training pipelines, our
%approach successfully trains regressors with highly sparse target
%distribution, is less sensitive to initial learning rate and trains
%faster. Furthermore, using the quantitative measures to estimate
%deformation stability of trained networks \cite{deformstab}, we
%observe that the networks trained using Adaloss are more stable.  We
%demonstrate the effectiveness of this method on a medical
%dataset of fluoroscopy scans (X-ray images) where task is to predict
%the catheter tip (single landmark regression) with high precision and
%300W \cite{300w}, a widely used benchmark dataset of 68 landmarks annotated on in-the-wild faces.

\section{Related work}

Availability of large-scale datasets \cite{flic, mpii} and powerful
GPUs allowed deep neural network based methods to outperform former methods on
landmarks detection. Toshev et al. were the first to reach
state of the art results with a DNN on the FLIC
dataset \cite{flic} with DeepPose \cite{deeppose}. Their approach
directly regress the coordinates of the body joints. Later
on, Thompson et al.~\cite{thompson} proposed to focus on regressing
heatmaps centered in the body joint locations instead of directly
predicting the coordinates.  This method outperformed direct
coordinate regression method especially on challenging images. The
intuition here was that directly regressing coordinates adds
unnecessary learning complexity by trying to map RGB to XY coordinates.
They also proposed to mix deep convolutional networks with
graphical models by jointly training a Markov Random Field to model the
correlation between the body keypoints.

Wei et al.~\cite{wei} proposed to use deeper convolutional networks
for directly regressing the heatmaps, by using multiple encoding-decoding
networks iteratively, with intermediate supervision. This method
was then improved by Newell et al.~\cite{hg} who extended the networks
by adding skip-connections between the convolution and deconvolution modules
to better combine the features across multiple scales. Their proposed stacked
hourglass architecture is as of today one of the most common architectures
for landmark localization networks. Dong et al.~\cite{sbr} recently proposed
a new method for improving temporal consistency for detecting facial landmarks
in videos by augmenting the training loss with a registration loss based on
optical flows. Honari et al.~\cite{honari2018} investigated the situation
where precise annotations are only provided for a small subset. They proposed
a multitasking framework which can leverage from imprecise annotations such as
regression and classification annotation to improve precise annotation such
as landmarks. Finally, Dong et al.~\cite{san} focused on augmenting the training
data by introducing a new Generative Adversarial Network (GAN) \cite{gan} to generate
a pool of style-aggregated face images, and showed state-of-the-art results on the
challenging 300-W face dataset.

While most recent works focused on improving model architectures and data augmentation,
our approach aims to propose a better objective function that may
be used in any heatmap regression setup and does not require problem
dependent parameters.

\section{Adaloss}

Adaloss, like many other deep-learning based landmark localization methods in the literature, represents the landmark locations as heatmaps. These heatmaps are created by placing multi-variate Gaussian distributions at the ground truth landmark locations. However, unlike existing methods, Adaloss does not use fixed target heatmaps throughout the training but instead  updates them as the training progresses. 

Adaloss begins the training with less precise (i.e. large variance) targets and iteratively makes them more precise (i.e. decrease the variance) during the training. This may be interpreted as progressively altering the difficulty of the task on hand and hence can be considered as a curriculum method. In the beginning, the network tries to solve an easier problem, and once that easy problem is solved, it moves on to another more difficult problem, but uses what is learned so far as a strong initialization.

However, to facilitate convergence during training, the curriculum must be appropriately defined. Here, we propose to use a simple additive update equation to set the variance for the target of each landmark, $l$ after each epoch, $t$ (note that, for keeping the notation simple, we are using ``standard deviation'' and $\sigma$, instead of ``variance'' and $\Sigma$ in our explanations, but depending on the application the target distributions may be one, two, or three dimensional):
\begin{equation}
\sigma^l_{t+1} = \sigma^l_{t} + \Delta\sigma^l_{t}
\label{eq:simple}
\end{equation}
where, $\Delta\sigma^l_{t}$ is the change in the standard deviation for the target of landmark $l$ in epoch $t$. Note that for multi-variate tasks, we simply employ isotropic targets, where each diagonal element of the target variance is defined as in Equation \ref{eq:simple}.

A naive approach would be to set $\Delta\sigma^l_{t} = -k$ as a constant decay parameter, linearly decreasing the standard deviation after each epoch. Even though this naive approach may work for some problems it would not generalize well to all problems. For some, the constant decay would be very fast and hence would affect the trainability and for others, it would be very slow making the learning very inefficient. This is also the case when training for regressing multiple landmarks at once. Each of these individual landmark regression problems may have a varying level of difficulty and hence a constant decay parameter would not work for all.

%%%%%%%%%%%%%%%%%%%%%%%%%%%%%%%

Thus $\Delta\sigma^l_{t}$ should instead be set according to what has been observed in the training so far. Motivated by how AdaDelta \cite{adadelta} method sets the learning rate, we compute $\Delta\sigma^l_{t}$ based on the previous losses in a fixed window of size $w$. Note that, unlike AdaDelta, we use the actual loss values within this training window, and hence store the complete loss history instead of a running average. The history of the losses is defined as:
\begin{equation}
H^l_t = [L^l_{t - i}], 0 \le i < w
\end{equation}

To characterize the evolution of the loss function, we compute the
loss variance $V^l_t$ which is set to be the variance of the values in
the loss history vector $H^l_t$:
\begin{equation}
V^l_t = \frac{1}{w}\left(\sum_{i=0}^{w-1} (L_{t-i}^l - \mathop{{}\mathbb{E}}[H^l_t])^2\right)
\end{equation}

If the loss variance is decreasing, it means that the loss function has not changed significantly during the last $w$ epochs. This suggests that the current landmark localization task (with the current standard deviation values) has converged or is close to convergence. In this case, it would help to decrease the standard deviation and attempt to solve a more difficult task. To this end, we compute the ratio of the previous variance $V^l_{t-1}$ with the current variance $V^l_{t}$ and use it as our standard deviation decay:
\begin{equation}
\Delta\sigma^l_{t} = 1 - \frac{V^l_{t-1}}{V^l_t}
\end{equation}

For numerical stability, we also introduce a regularizer $\rho$ and finally, putting it all together, we obtain the following update equation for the target standard deviations:
\begin{equation}
\sigma^l_{t+1} = \sigma^l_{t} + \rho \cdot \left(1 - \frac{V^l_{t-1}}{V^l_t}\right)
\label{eq:final}
\end{equation}
where we set $\sigma^l_{0}$ to be the quarter of the target resolution. This provides a good starting point as it addresses the sample bias problem
associated with standard heaptmap regression methods.

%%%%%%%%%%%%%%%%%%%%%%%%%%%%%%%

Note that in Equation \ref{eq:final} an increase in the loss variance would cause a corresponding increase in the target standard deviation. On one hand this may help escaping from a local minimum, but on the other hand it could possibly lead to divergence with an overly relaxed regression task. Thus, in practice, we take into consideration the training loss in determining whether to apply a \textit{non-increasing update restriction rule} or not. 

This especially plays an important role while training multiple landmarks. Since the landmarks may be of varying difficulty, $\sigma^l$ for one landmark may reduce faster than others. Without the non-increasing update rule, we could have situations where the training is stuck in an oscillatory behavior. However, the network might need to compromise on the precision of some landmarks during the optimization process. In this case, the network may choose to focus on a particular set of landmarks to the detriment of the others, for which the loss may stagnate or increase. It might thus be useful to allow Adaloss to take a step back and increase the standard deviation for a given landmark \textit{when the loss variance increases together with the training loss}. 

%An important with any update parameter like $\sigma$ or even learning
%rate is the initialization. To this end, we set the $\sigma$ to a
%quarter of the heatmap resolution as such as heatmap would have an
%equiprobable distribution of zeros an non-zeros values, preventing the
%network from favoring zero pixel values (due to sample bias). This use
%of this value to initialize $\sigma$ is also validated in all our
%experiments.

%\begin{algorithm}[t]
%  \caption{Updating the standard deviation of one landmark}\label{algo}
%  \begin{algorithmic}[1]
%    \Require $H$, the history of the last $w$ losses.
%
%    $V_{t-1}$, the last computed variance ($V_0 = 0$).
%
%    $\sigma_{t-1}$, the last standard deviation ($\sigma_0 = 0.25 * d$, where $d$ is the output resolution).
%
%    $\rho$, the regularizer.
%
%    \State $V_t \gets Var(H)$
%    \State $\Delta_v \gets \frac{V_{t-1}}{V_t}$
%    \State $x \gets \sigma_{t
%      -1} + \rho (1 - \Delta_v)$
%    \If{$x < \sigma_{t-1}$ and $x \ge 1$}
%    \State $\sigma_t \gets x$
%    \EndIf
%    \State $V_{t-1} \gets V_t$
%\end{algorithmic}
%\end{algorithm}

\section{Experiments}

We evaluated our method on three very different landmark localization problems: catheter tip detection in X-ray images (single instance of a single landmark that needs to be localized very precisely), facial feature localization (single instances of multiple landmarks where the achievable precision is landmark-dependent), and surgical instrument detection in endoscopy images (multiple instances of multiple landmarks that needs to be localized with minimal false-positives).

\subsection{Technical details}

For all experiments, we used popular network architectures from respective
domains and ensured the initial parameters
are set appropriately when comparing with and without Adaloss. The
models were trained and validated using PyTorch
\cite{paszke2017automatic}. All our experiments used the same following
Adaloss parameters, $w=3$, $\rho=0.9$ and $\sigma_0$ set to a quarter of the image resolution.
\par
For the catheter tip detection task, we used a U-Net architecture
\cite{unet} with 9 convolutional layers with a fixed kernel size of $5$ using Batch Normalization \cite{batchnorm} and Rectified Linear Unit (ReLU) \cite{relu}. 
We trained the network for 100 epochs.
\par
For the face landmark detection task, we used a Dense U-Net architecture
\cite{denseunet} with 23 convolutional layers with a fixed kernel size of $11$ using Batch Normalization and ReLU. We trained the network for 150 epochs.
\par
For the endoscopy instruments detection, we used a U-Net architecture
\cite{unet} with 9 convolutional layers using Batch Normalization and Leaky ReLU. 
We trained the network for 300 epochs.

\subsection{Single instance of a single landmark}

\begin{figure}[t]
  \centering
  \includegraphics[width=0.9\linewidth]{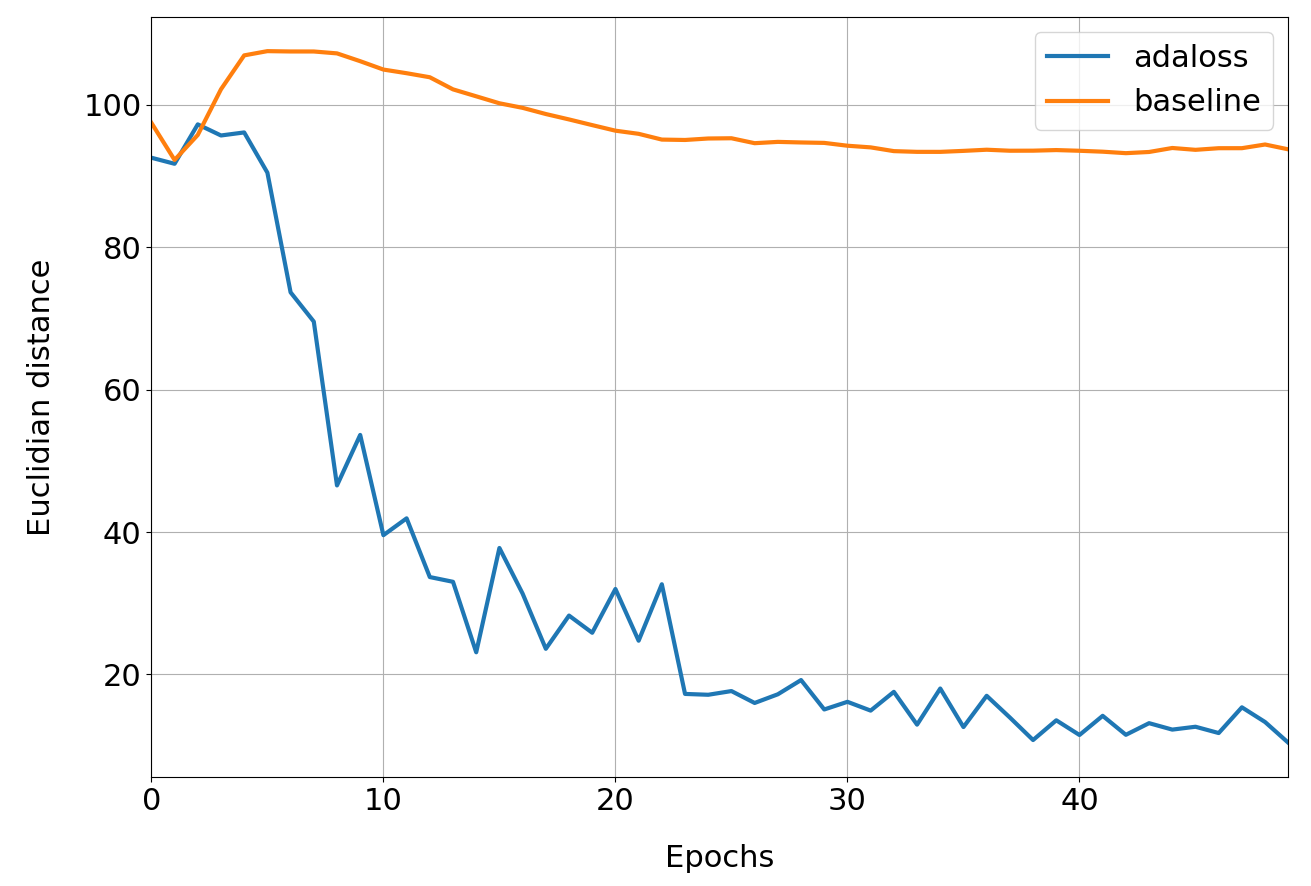}
    \caption{Evolution of the euclidian distance on the
    CathDet validation set using Adadelta with and without
    Adaloss.
  }
  \label{fig:comp_adadelta}
\end{figure}

\begin{figure*}[ht]
  \begin{center}
    \centering
	\includegraphics[width=0.9\textwidth]{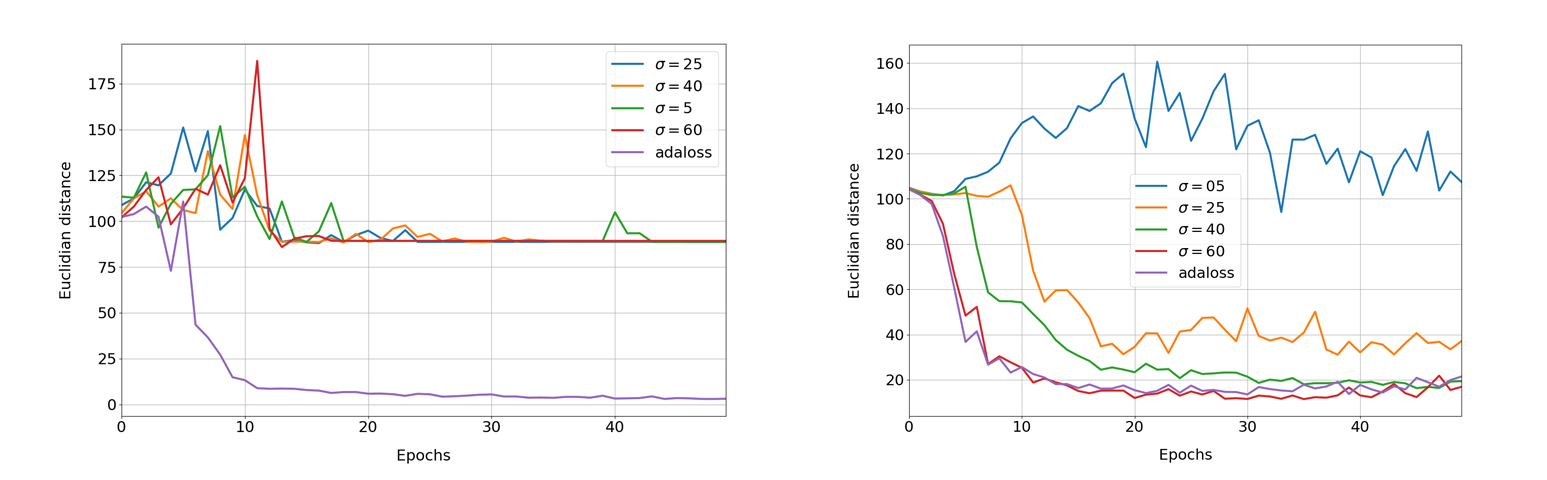}
	\caption{Evolution of the Euclidian distance on the
          CathDet validation set for multiple fixed values of sigma
          and Adaloss. Left figure is using the Adam optimizer with a learning
          rate of $10^{-3}$ and on the right using a learning rate of $10^{-4}$.
          While the models with fixed sigmas can only train with the lower
          learning rate, models trained with Adaloss trains faster and better
          with higher learning rate.}
	\label{fig:comp_sigma_cathdet}
  \end{center}
\end{figure*}

Our first experiment is conducted on a medical dataset of 3300
fluoroscopy scans (X-ray images) annotated with a single landmark
which marks the tip of a catheter\footnote{Due to compliance reasons, this
dataset is not publicly available.}. We refer to this dataset as ``Cathdet''.
3000 images are used for training and 300 are used for testing. The output of the network
is a heatmap with resolution of 256x256.

%Detecting the catheter tip is the first
%step in several image-guided interventional procedures
%\cite{cathtip18}. Often the desired precision of this marker is very
%high and hence we report the accuracy of our models using the
%Euclidian distance to the ground truth location. 

%Our first experiment consists in detecting the tip of the catheter 
%on medical x-ray images, which defines a very precise position.
%The final target is a probability distribution with very small variance,
%which is difficult to train, and often leads to
%networks predicting all black images. 
We first examined the benefits of Adaloss in terms of training stability. As setting the learning rate
also known to have a significant impact on training stability, a natural
comparison to our method is to use an adaptive learning rate method such as Adadelta
\cite{adadelta}. However, as illustrated in figure \autoref{fig:comp_adadelta}, only adapting the learning rate was not sufficient for handling problems with significant sample bias. 

In addition to using a smaller
learning rate, another possibility could be to increase
sigma. However, finding the appropriate value for sigma highly depends
on the problem and while using higher sigma certainly helps to stabilize the
training, it also increase tolerance to errors in prediction and ends
up altering the final accuracy.

We compared these approaches to Adaloss and trained multiple models
with different learning rates and different values for sigma
(\autoref{fig:comp_sigma_cathdet}). Here, when using the Adam
optimizer with a learning rate of $10^{-3}$, models with fixed sigmas
are not able to train, even with high values such as $\sigma=40$.
When decreasing the learning rate by a factor of 10, models with fixed
sigmas no longer diverge, at the cost of slower training and lower
accuracy (except for the one with $\sigma=5$ which still could not
train). In comparison, models trained with Adaloss can train with both
learning rates, and converges faster and better with the higher
learning rate. The best non-adaptive method here is using
$\sigma=40$. The network which is trained with $\sigma=60$ starts
with a better accuracy, but the error starts to increase after a certain
number of epochs, due to the excessive tolerance.
\autoref{table:comp_acc_cathdet} shows best results on testing set for
all values of sigmas and Adaloss after 150 epochs. Model trained with
Adaloss shows the lowest \textbf{1.19} Euclidian distance.

\begin{center}
  \begin{table}[t]
    \centering
  \begin{tabular}{|l|l|l|l|l|l|}
    \hline
    & $\sigma=5$ & $\sigma=25$ & $\sigma=40$ & $\sigma=60$ & Adaloss\\
    \hline
    Dist. & 87.05 & 19.27 & 8.71 & 11.46 & \textbf{1.19}\\
    \hline
  \end{tabular}
  \caption{Euclidian distance on CathDet testing set for multiple values
  of sigma compared to Adaloss.}
  \label{table:comp_acc_cathdet}
  \end{table}
\end{center}
\begin{figure}[h!]
  \begin{center}
	\includegraphics[width=\linewidth]{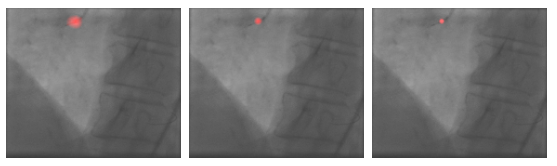}
	\caption{Example of a challenging sample from the validation
          set. In the early training (left image), the proposed method
          generates targets with a large variance. At convergence (middle image), target variance
          becomes much smaller and precisely defines the tip of the catheter.
          Right image presents the ground truth catheter tip position.}
	\label{fig:comp_preds}
  \end{center}
\end{figure}

\begin{figure}[h]
  \begin{center}
    \centering
	\includegraphics[width=0.9\linewidth]{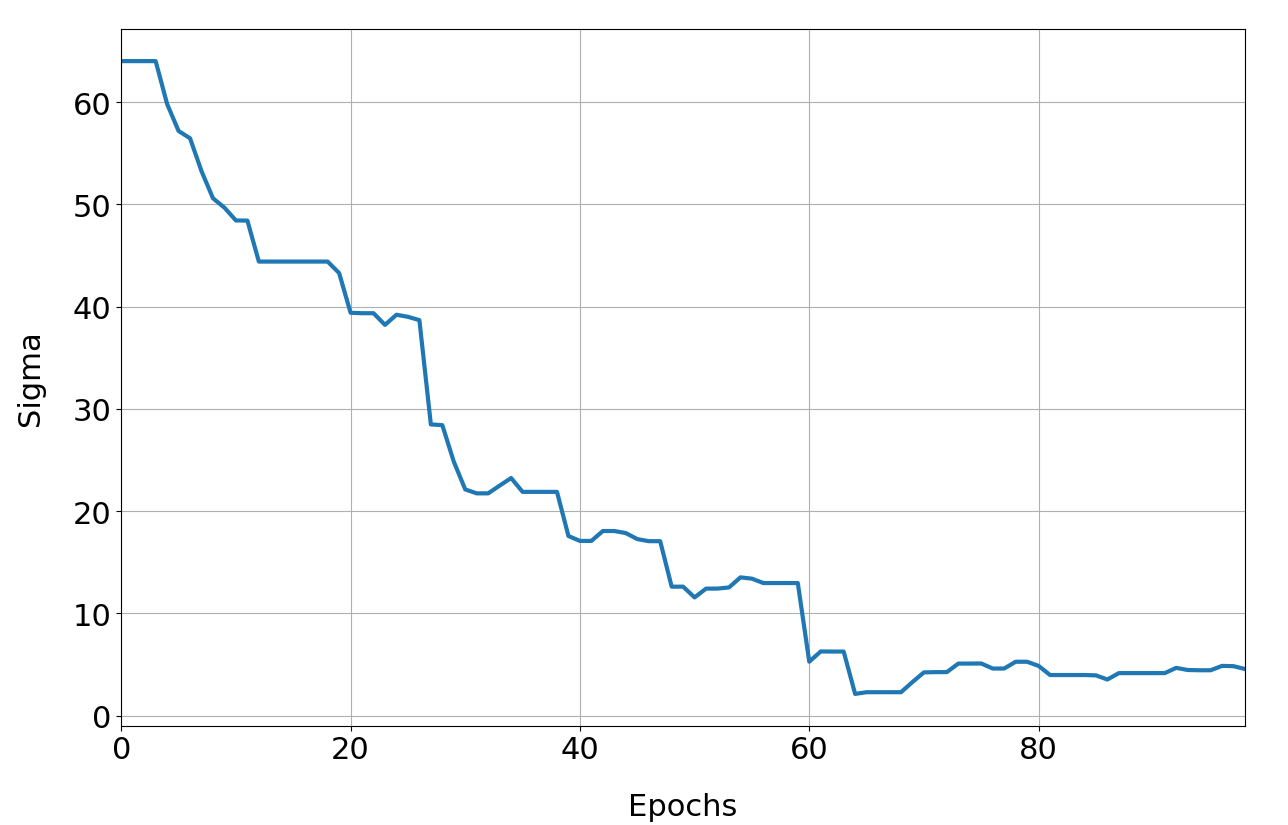}
	\caption{Evolution of the standard deviation during the training of the
	catheter tip detection model. At convergence, sigma stops decreasing, 
	showing that it has reached the optimal value (Here around 5 pixels).
	This final value may represent the variance in the annotation.}
	\label{fig:sigma_cathdet}
  \end{center}
\end{figure}

Finally, we investigated the evolution of the standard deviation across 
the training (\autoref{fig:sigma_cathdet}). After 60 epochs, the model
converges and the standard deviation stops decreasing. This shows that
Adaloss has found the optimal value for sigma for this problem. Here,
the value of the final standard deviation is about \textbf{5 pixels}. We
found this value to be representative of the variance in the ground
truth annotations of the tip of the catheter. The variance here is
very small as the target defines a precise point on the image.

\subsection*{Exploring the kernel smoothness}

Recently, Ruderman et al.~\cite{deformstab} explored how filter
smoothness is an important factor for determining the stability of
convolutional networks to deformation. They presented a method for
measuring filter smoothness using Normalized Total Variation (NTV) and
show how this metric correlates with the ability of the network to
learn the deformation bias. We computed this metric on best networks
with and without Adaloss and reported the NTV
scores in \autoref{table:ksmooth}. Model trained with Adaloss presents
a much smaller NTV (\textbf{2.80} vs 3.03),
indicating a potential impact of using Adaloss on filter smoothness.

\begin{table}[h]
  \centering
  \begin{tabular}{|l|l|l|}
    \hline
    & Fixed sigma & Adaloss \\
    \hline
    All & 3.03 & \textbf{2.80} \\
    \hline
    Last & 1.83 & \textbf{1.68} \\
    \hline
  \end{tabular}
  \caption{Normalized Total Variation (NTV) score
    with and without Adaloss. `all' represents
    the aggregate NTV on all the layers, and `last' the NTV on the final
    convolutional layer. Model trained using Adaloss has smoother
    filters, indicating potential improvement in network stability.}
  \label{table:ksmooth}
\end{table}

\subsection{Single instances of multiple landmarks}

We conducted our second experiment on the 300-W dataset \cite{300w,300w_2,300w_3}.
This dataset is an aggregation of five face datasets annotated with 68
landmarks: LFPW \cite{lfpw}, AFW \cite{afw}, HELEN \cite{helen}, XM2VTS and IBUG. It is often used as a benchmark dataset for evaluating facial feature localization methods \cite{poseinvar, MDM, CFSS}. We followed the same splitting as in \cite{san, sbr, CFSS, twostage}, and used all training samples
from LFPW and HELEN and the whole AFW dataset for training. Testing
samples from LFPW and HELEN are used as the ``common'' testing set,
while IBUG testing set is considered as ``challenging''. This
splitting results in a training set of 3148 images, a common testing
set of 554 images and a challenging set of 135 images. The union of
the common and the challenging set is used as the ``full'' testing
set.

When working on multi-landmarks problems, the difficulty resides
in the difference between the landmarks. For a given set of
landmarks, some of them might have well defined features and a small
variance in the annotation (e.g. eye landmarks) while some
others might have varying features (this is the case for the mouth,
which could be either open or closed), or may have a large variance in
the annotation as for the jaw which defines a broad region of the face.
%%(\autoref{fig:var_annot}).

%\begin{figure}[h]
%  \begin{center}
%    \centering
%	\includegraphics[width=0.9\linewidth]{var_annot}
%	\caption{Variance of each landmark point with regard to three expert human annotators provided in \cite{300w}.}
%	\label{fig:var_annot}
%  \end{center}
%\end{figure}

Adaloss addresses this issue by using different standard deviations
for different landmarks. Here, the update rule is the same as for the
single landmark case, except that the loss is computed per landmark,
meaning that at a given step in the training, one landmark can see its
standard deviation decrease while the other ones might not. To further
increase the impact of having varying target distributions, we normalize
each target heatmap with its mean.

\begin{figure}[h]
  \centering
  \includegraphics[width=\linewidth]{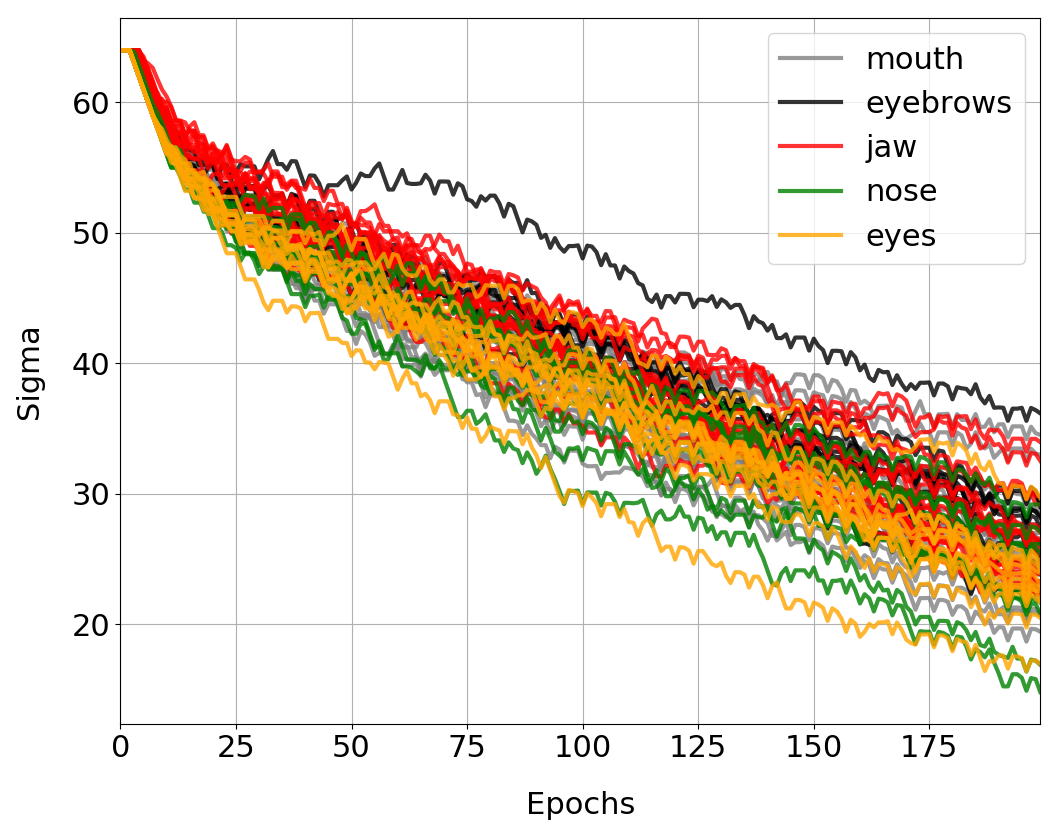}
  \caption{Evolution of the standard deviation for landmarks in
    different regions. Well defined regions such as eyes and nose
    decrease fast while locations subject to occlusions or variance
    in annotations decrease slower.}
  \label{fig:sigma_var_face}
\end{figure}

\autoref{fig:sigma_var_face} shows how Adaloss evolves during
training. We aggregated the evolution of the standard deviation for
different regions. Jaw and eyebrows are difficult landmarks: jaw
landmarks suffer from significant variance in the annotations and
eyebrows landmarks are very sensitive to face orientation. It is
interesting to see that the standard deviation decreased slowly for
these landmarks and kept a relatively high standard deviation at
convergence. On the other hand, the standard deviation decreased faster
for the mouth landmarks, and even faster and lower for the eyes and
nose landmarks, which are less subject to variance in annotations.

These observations suggest that Adaloss was indeed able
to adapt the standard deviation for the landmarks with regard to their
difficulty.

\begin{figure*}[t]
  \begin{center}
    \centering
	\includegraphics[width=\textwidth]{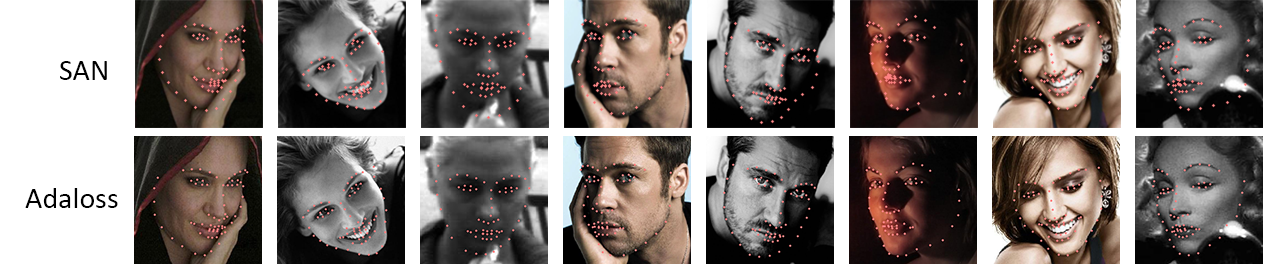}
	\caption{Example of difficult samples from the challenging
          dataset. Our method is more robust to occlusions and better
          captures local features even at low resolution, or in bad
          lighting conditions.}
	\label{fig:quali_face}
  \end{center}
\end{figure*}

We trained our Dense U-Net end-to-end minimizing the Mean Squared Error
(MSE) using the Adam optimizer \cite{kingma2015} with an initial learning
rate of $10^{-4}$. We augmented the training set using rotations from
$-30$ to $+30$ degrees. We used the ground truth bounding box to prepare
the input for our network at a resolution of $256$x$256$. \autoref{table:comp_nme} compares the
normalized mean errors (NME) with recent state-of-the-art methods.
Following these works, we use the inter-ocular distance (IOD) to normalize
the mean error. Our network trained with Adaloss shows the best results
on all testing sets. On the common test set, our approach outperforms
the previous best method with a NME of \textbf{2.76} vs $3.28$.
On the challenging set, our method reached a NME of \textbf{5.61} vs
$6.60$. Overall NME on the full test is \textbf{3.31} vs $3.98$.
We also compare our approach with the widely used Dlib \cite{dlib},
which was also initialized using the provided ground truth bounding box
from 300W. Again, our approach presents better results on all testing
sets, and especially on the challenging one (\textbf{5.61} vs
$19.39$), showing the robustness of our approach.
\autoref{fig:quali_face} shows qualitative results from our
method compared to \cite{san} on difficult samples from the challenging
set. Results from \cite{san} were generated using provided code
and models. Our method was able to better address the occlusion
issues and learned a better model of the face.

\begin{table}[h]
  \centering
  \begin{tabular}{|l|l|l|l|}
    \hline
    Method & Common & Challenging & Full \\
    \hline
    LBF \cite{LBF} & 4.95 & 11.98 & 6.32 \\
    CFSS \cite{CFSS} & 4.73 & 9.98 & 5.76 \\
    MDM \cite{MDM} & 4.83 & 10.14 & 5.88 \\
    TCDCN \cite{TCDCN} & 4.80 & 8.60 & 5.54 \\
    Two-Stage \cite{twostage} & 4.36 & 7.42 & 4.96 \\
    RDR \cite{RDR} & 5.03 & 8.95 & 5.80 \\
    Pose-Invariant \cite{poseinvar} & 5.43 & 9.88 & 6.30 \\
    RCN \cite{honari2018} & 4.20 & 7.78 & 4.90 \\
    SAN \cite{san} & 3.34 & 6.60 & 3.98 \\
    CPM + SBR \cite{sbr} & 3.28 & 7.58 & 4.10 \\
    \hline
    Dlib \cite{dlib} & 4.38 & 19.39 & 7.32 \\
    \hline
    Adaloss & \textbf{2.76} & \textbf{5.61} & \textbf{3.31} \\
    \hline
  \end{tabular}
  \caption{Normalized mean errors (NME) on the 300-W dataset.
    Proposed approach shows best results on all testing sets.}
  \label{table:comp_nme}
\end{table}

\subsection{Multiple instances of multiple landmarks}

For our final experiment, we used the MICCAI 2015 challenge EndoVis dataset \cite{endovis}, a high resolution ($576$ x $720$) intra-operative dataset of laparoscopic images with annotated
instruments \cite{tmi}.

\begin{figure}[H]
  \begin{center}
    \centering
	\includegraphics[width=0.7\linewidth]{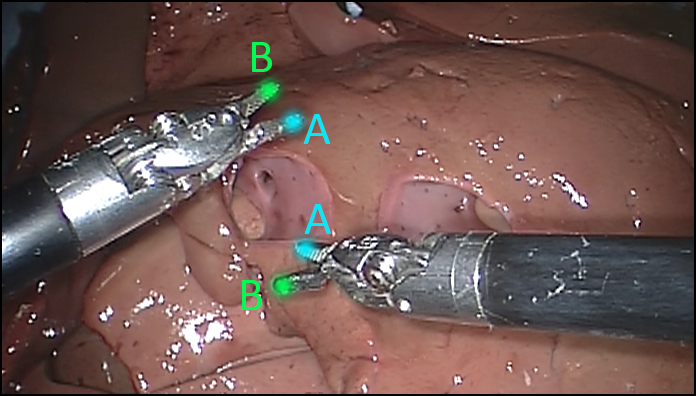}
	\caption{Example of multiple instances of a \textit{Needle Driver} from the testing set.
          Landmarks (A) define the \textit{Right Clasper} and landmarks (B)
          the \textit{Left Clasper}.}
	\label{fig:endovis_example}
  \end{center}
\end{figure}

In this experiment, we focused on only two landmarks: ``Right Clasper''
and ``Left Clasper'', which correspond to the extremities of the tool
(\autoref{fig:endovis_example}). These landmarks are the most important to track as
they define the parts of the tool that are interacting with the tissues. Here again,
landmarks are represented with Gaussian heatmaps, but in this case,
there can be multiple Gaussians in each target heatmap, since multiple instances of a particular tool type may appear in the frames simultaneously.

When detecting multiple instances of a landmark on a given image, it
is not possible to use argmax anymore to get the final coordinates.
We first applied a median filter with a kernel dimension of [$7$, $7$]
followed by non-maxima suppression (NMS) using a window size of $10$
pixels.  Finally, the centers of the remaining clusters of points
correspond to the localization of the detected landmarks.

\begin{figure*}[ht]
  \begin{center}
    \centering
	\includegraphics[width=0.9\textwidth]{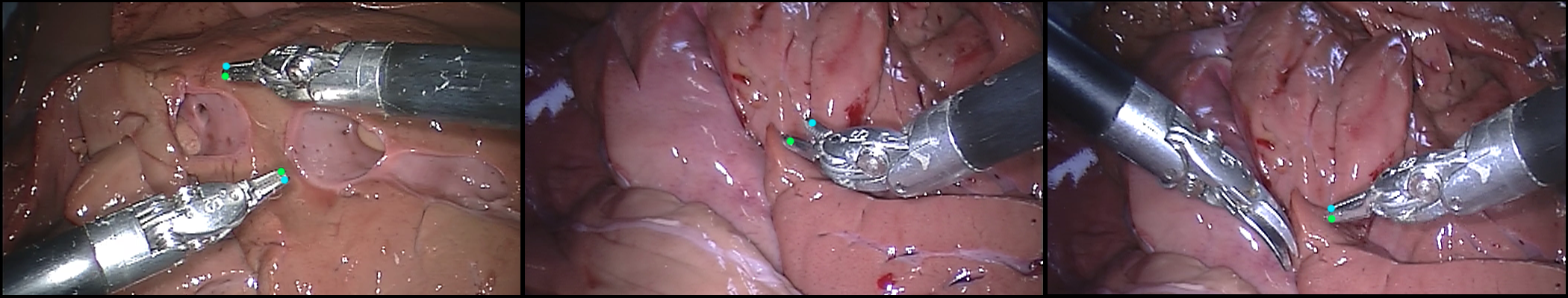}
	\caption{Results from the proposed approach on both EndoVis testing set.
          Left and center images show results from the testing set without Scissors.
          Right image shows a result from the full testing set where the
          \textit{Curved Scissor} was not misinterpreted as a \textit{Needle Driver}.}
	\label{fig:quali_endovis}
  \end{center}
\end{figure*}

The EndoVis dataset provides
images of two different tools: ``Curved Scissor'' and ``Needle
Driver''. Current state-of-the-art method for detecting instruments in
endoscopic images uses a multi-stage approach. It firsts
locates joints and association between joints using a detection
network, then it refines these joints through a regression network and then
retrieves the final location using maximum bipartite graph matching
\cite{tmi}. While this approach performs landmark detection on
multiple instruments with no distinction, it is often desirable to
distinguish one instrument from another.  

We trained our network end-to-end minimizing the Mean Squared
Error (MSE) using the Adam optimizer \cite{kingma2015} with an initial
learning rate of $10^{-3}$. 
We computed Precision, Recall and Euclidian distance of our predictions
on the testing set of 910 images. This testing set is composed of images with both
curved scissors and needle drivers, for which there may be multiple instances in the
same frame. Results can be seen in \autoref{table:endovis}. Our model performs a
left clasper precision of $\textbf{90.50\%}$ with a recall of $\textbf{95.56\%}$ on the
needle driver. In comparison, the multi-stage approach from \cite{tmi} performs
a left clasper precision of $86.65\%$ with a recall of $86.28\%$ on both tools with no
distinction.

One important thing to note is that the curved scissor is not present in the training set.
Our end-to-end network was thus trained to detect only needle drivers. To investigate
the robustness of our network and its ability to distinguish tools, we evaluated
our approach on two subsets, one composed of 308 images with needle driver only, and other
composed of 301 samples where each image contains both needle drivers and curved scissors.
On the first subset, our model detected the claspers with a mean precision of $96.56\%$
across all landmarks. On the second subset, the mean precision dropped to $92.16\%$. While
this decrease in precision is expected due to the presence of unseen tools, it is interesting
to see that the mean precision is still over $90\%$, showing that our model was able to learn
discriminative features to distinguish different instruments. Qualitative results from
these subsets can be found in \autoref{fig:quali_endovis}.

\begin{table}[h]
  \centering
  \begin{tabular}{|l|c|c|}
    %% \hline
    %% & Recall (\%) / Precision (\%) / Euclidian distance (px) & \\
    \hline
    & Left Clasper & Right Clasper \\
    \hline
    No Scissor & $93.18$ / $98.86$ / $7.23$ & $96.59$ / $94.26$ / $6.38$ \\
    \hline
    With Scissors & $98.02$ / $97.10$ / $7.21$ & $97.37$ / $87.23$ / $6.56$ \\
    \hline
    Full set & $95.56$ / $90.50$ / $6.83$ & $96.38$ / $81.20$ / $6.61$ \\
    \hline
  \end{tabular}
  \caption{Quantitative results of the Adaloss model on different testing
    set of EndoVis: Recall (\%) / Precision (\%) / Euclidian distance (px).}
  \label{table:endovis}
\end{table}

\section{Conclusion}
%% We have presented a novel approach to train deep networks for
%% regressing landmarks in images and validated the method on multiple
%% problems. Using adaloss, we have shown that can be trained to detect
%% landmarks with higher accuracy and precision compared to previous
%% methods. Based on our experiments, we expect adaloss to be applicable
%% to problems beyond landmark detection such as segmenting vessels in
%% X-ray images (long/thin images structures) which are otherwise
%% difficult to segment.

We presented a novel, application-independent approach, ``Adaloss'', for training deep neural networks for landmark localization. Our method systematically increases the problem difficulty during training and as a result implicitly creates a \textit{training curriculum}. It addresses the sample bias problem in training, which is arguably the main limitation of the existing heatmap regression methods. We demonstrated the effectiveness of our method on three challenging domains: detecting catheter tips in X-ray images, localizing surgical instruments in endoscopy images, and facial feature localization on in-the-wild images, where we significantly advance the state-of-the-art. While progressively updating the objective
function for landmark regression appears to be effective, the
viability of such a method has not been explored for other regression
or classification tasks and is yet open to discussion.

\bibliographystyle{unsrt}
\bibliography{cvpr_2019}

\end{document}